\documentclass[10pt,twocolumn,twoside]{IEEEtran}
\IEEEoverridecommandlockouts
\usepackage{caption}
\usepackage{subfigure}
\usepackage{cite}
\usepackage{amsmath,amssymb,amsfonts}
\usepackage{algorithmic}
\usepackage{graphicx}
\usepackage{textcomp}
\usepackage{xcolor}
\usepackage{cite}
\usepackage{amsmath,amssymb,amsfonts}
\usepackage{algorithmic}
\usepackage{graphicx}
\usepackage{textcomp}
\usepackage{xcolor}
\usepackage{algorithm}
\usepackage{comment}
\usepackage{setspace}
\usepackage{color}
\usepackage{booktabs}
\usepackage{multirow}
\usepackage{epstopdf}
\usepackage{amsmath}
\usepackage{amsthm}
\usepackage[margin=0.7in]{geometry}

\def\BibTeX{{\rm B\kern-.05em{\sc i\kern-.025em b}\kern-.08em
    T\kern-.1667em\lower.7ex\hbox{E}\kern-.125emX}}
\begin{document}

\title{Adversarial Learning for Implicit Semantic-Aware Communications}
\author{\IEEEauthorblockA{Zhimin Lu\IEEEauthorrefmark{1}, Yong~Xiao\IEEEauthorrefmark{1}\IEEEauthorrefmark{2}\IEEEauthorrefmark{5}, Zijian Sun\IEEEauthorrefmark{1}, Yingyu Li\IEEEauthorrefmark{3},  Guangming~Shi\IEEEauthorrefmark{2}\IEEEauthorrefmark{4}\IEEEauthorrefmark{5}, Xianfu Chen\IEEEauthorrefmark{6}, Mehdi Bennis\IEEEauthorrefmark{7}, and H. Vincent Poor\IEEEauthorrefmark{8} \\
\IEEEauthorblockA{\IEEEauthorrefmark{1}School of Elect. Inform. \& Commun., Huazhong Univ. of Science \& Technology, Wuhan, China}\\
\IEEEauthorblockA{\IEEEauthorrefmark{2}Peng Cheng Laboratory, Shenzhen, China}\\
\IEEEauthorblockA{\IEEEauthorrefmark{3}School of Mech. Eng. and Elect. Inform., China Univ. of Geosciences, Wuhan, China}\\
\IEEEauthorblockA{\IEEEauthorrefmark{4}School of Artificial Intelligence, Xidian University, Xi'an, China}\\
\IEEEauthorblockA{\IEEEauthorrefmark{5}Pazhou Laboratory (Huangpu), Guangzhou, China}\\
\IEEEauthorblockA{\IEEEauthorrefmark{6}VTT Technical Research Centre of Finland, Finland}\\
\IEEEauthorblockA{\IEEEauthorrefmark{7}University of Oulu, Oulu, Finland}\\
\IEEEauthorblockA{\IEEEauthorrefmark{8}Department of Elect. \& Computer Eng., Princeton University, Princeton, NJ, USA}\\
}
\thanks{*This work has been accepted at the Proceedings of the IEEE International Conference on Communications (ICC) conference, Rome, Italy, May 2023. Copyright may be transferred without notice, after which this version may no longer be accessible.}
}
\maketitle

\begin{abstract}
Semantic communication is a novel communication paradigm that focuses on recognizing and delivering the desired meaning of messages to the destination users. Most existing works in this area focus on delivering explicit semantics, labels or signal features that can be directly identified from the source signals. In this paper, we consider the implicit semantic communication problem in which hidden relations and closely related semantic terms that cannot be recognized from the source signals need to also be delivered to the destination user. We develop a novel adversarial learning-based implicit semantic-aware communication (iSAC) architecture in which the source user, instead of maximizing the total amount of information transmitted to the channel, aims to help the recipient learn an inference rule that can automatically generate implicit semantics based on limited clue information. We prove that by applying iSAC, the destination user can always learn an inference rule that matches the true inference rule of the source messages. Experimental results show that the proposed iSAC can offer up to a 19.69 dB improvement over existing non-inferential communication solutions, in terms of symbol error rate at the destination user.   
\end{abstract}

\vspace{-0.1in}

\section{Introduction}

Semantic communication has recently been attracting significant interest, driven mostly by the recent surge in the demand of data-hungry and resource-consuming smart and human-oriented services, such as Tactile Internet\cite{XY2018TactileInternet}, intelligent transportation systems\cite{XY2022AdaptiveFog}, eXtended Reality (XR), and digital twins. Different from the traditional content-agnostic communication solutions that focus on transmitting data packets from one user to another, while ignoring the semantics, semantic communication focuses on sensing, recognizing, utilizing, and delivering  the key meaning of transported messages throughout the network. Recent results have demonstrated that semantic communication has the potential in significantly improving communication efficiency, quality-of-experience (QoE) of various services, and imbuing more high-level human-like capabilities into communication networks\cite{shi2021semantic}.     


The concept of semantic communication was first introduced in 1949, where Weaver defined three levels of communication problems\cite{weaver1949recent}. The Shannon theory has been considered as the solution for level 1 problem, also called the technical problem of communications. The semantic and effective communication problems have been defined as level 2 and level 3 problems which address delivering ``the desired meaning" and influencing the conduct of the destination user in the desired way. These definitions have attracted significant interest in extending Shannon theory to investigate the semantic and effective communication problems. Carnap et al. \cite{carnap1952outline} observed that there exists a fundamental paradox, called the Bar-Hillel-Carpnap (BHC) paradox, when applying Shannon theory to solve the semantic communication problems. In the BHC paradox, semantically incorrect statements, especially those contradicting with common-sense knowledge can always maximize the Shannon information due to their rarity.

Recently, there has been a growing interest in applying deep learning (DL) algorithms to improve the performance of communication networks. This has sparked many works to convert the problem of semantic communication into pattern recognition and/or classification problems that can be solved by DL-based algorithms. In these works, manually labeled objects, terms, and/or signal features that can be directly identified from various forms of signals are defined as semantics. For example, DL algorithms have been applied to identify semantics from text, image, and voice\cite{Santhanavijayan2021ASR} for various downstream communication tasks.

Recent observations suggest that information semantics can be much more than just the object labels. In fact, according to the original definition of semantics first introduced by Breal in 1897, semantics are the ``relationship between words and the knowledge they signify". Recent work in cognitive neuroscience also emphasizes that human users are able to express and infer complex implicit semantics by automatically inferring complex hidden relations among concepts and ideas. This motivates the work of this paper to investigate the implicit semantic communication problem. We define semantics that can be directly identified from the source messages as {\em explicit semantics} and focus on developing solutions to infer the implicit semantics, including hidden relations and relevant semantic concepts that cannot be identified from the source messages, by a destination user. More specifically, we introduce a representation model of the implicit semantic information and develop an adversarial learning-based implicit semantic-aware communication (iSAC) architecture in which the source user, instead of maximizing the information transmitted across the channel, tries to assist the destination user to learn an inference rule that can automatically infer implicit semantics based on the received clue information, e.g., explicit semantics, sent by the source user. We prove that by applying iSAC, the destination user can always learn an inference rule that matches the true inference rule of the source messages. Also, since the true inference rule of the source messages generated by human users can always be assumed to be semantically correct, the BHC paradox can be naturally solved. We conduct extensive simulations based on real-world datasets. Our results show that the proposed iSAC can achieve up to 19.69 dB improvement over existing non-inferential communication solutions, in terms of symbol error rate at the destination user. 

\section{Related Works}


Existing works in semantic communication can be roughly divided into three categorizes: information theory-based, machine learning-based, and cognitive neuroscience-inspired works. In particular, Carnap et al.\cite{carnap1952outline} elaborated semantic information measurements analogous to the binary-symbol-based information measurements in Shannon theory. Moreover, Bao et al.\cite{Bao2011TowardsTheorySemanticComm} further investigated the quantification of semantic information as well as semantic coding, and obtained initial results showing the feasibility of data compression and reliable communication from a semantic perspective. Motivated by the fact that semantic information can be learned and evolved through interaction, in our recent work\cite{xiao2022RateDistortion}, we have proposed a strategic semantic communication framework by combining game theoretic models with rate-distortion theory.

Recently, the powerful capabilities of machine learning, especially the DNNs-based learning algorithms, in pattern recognition and data classification have been introduced to solve the semantic communication problem. Most existing works have focused on the representation, identification, computation, and communication of explicit semantic, such as human-assigned labels and sample-related features or classes, that can be directly recognized from the transmitted messages. For example, the semantic communication systems for text transmission were developed in \cite{Guler2018SCGame}, while the semantic error was measured at the word level and sentence level, respectively. 
A lite distributed semantic communication system, named L-DeepSC, was proposed in \cite{Xie2021LDeepSC} also for text transmission.
For image transmission, Huang et al. \cite{Huang2021ImageSamanticCoding} adopted a GAN-based image semantic coding method for sending and reconstructing images in extreme low bit rate using the proposed semantic communication system. 
 For transmitting speech signals, an attention mechanism based semantic communication system was developed in \cite{Weng2021DeepSCS}, which is capable of identifying the essential information of speech signals under dynamic channel environments in telephone systems. Furthermore,  Seo et al. proposed a stochastic model of semantics-native communication (SNC) for generic tasks in \cite{seo2021semantics} infused with contextual reasoning to cope with the situation where the semantics vary over time and in different contexts. 

Inspired by the recent study in cognitive neuroscience suggesting that the human user is able to infer complex implicit semantics based on a limited clue/explicit information, we have investigated the implicit semantic communication in our recent works\cite{xiao2022ReasoningOnTheAir,XY2023CollaberativeJSAC,xiao2022RateDistortion, Liang2022Lifelong}. We have derived an information theoretic bound of the implicit semantic communication channel based on the rate distortion theory in \cite{xiao2022RateDistortion}. By formulating the implicit semantic reasoning process of the source user as a reinforcement learning process, an imitation learning-based solution has been proposed in \cite{XY2023CollaberativeJSAC} for the destination user to estimate the reasoning policy of the source user. One of the key issues of this work is that the proposed imitation learning-based solution always infers all the possible reasoning paths based on a maximum causal entropy framework. Also, the state space in the formulated reasoning policy increases in an exponential scale with the path length which may result in slow computation and reduce accuracy under certain scenarios.  
In this paper, we introduce a simple adversarial learning-based solution, iSAC, that allows the destination user to directly learn a much simplified approximated inference rule which will always output the most likely implicit term. Our proposed iSAC requires much less computational load and can deliver comparable performance to our previously proposed solution in many practical scenarios. 


\section{A General Semantic Communication Model}
\subsection{Representation of Implicit Semantic Information}
As mentioned earlier,
the semantics of a message should include the implicit relations that link the explicit semantic terms, e.g., the concepts and/or labels directly observed in the source messages, to the relevant implicit meanings. Therefore, in this paper, we define the semantic information of a given message as a tuple $\omega = \langle v, u_v \rangle$ where $v$ is a set of explicit semantic terms, e.g., labels or features, that can be directly recognized from the message, $u_v$ consists of the implicit relations and the connected semantic terms that are closely related to the explicit semantics $v$. $u_v$ cannot be directly observed from the source message but will need to be inferred based on the previous communications and inference preference of the source user. Let $\mathcal{V}$ be the set of all possible explicit semantic terms that can be expressed by the source user. We assume relations are undirectional and each implicit semantic term in $u_v$
is linked to a term in $v$ by a specific relation. We can therefore abuse the notation and use $u_v$ to denote a set of implicit semantic terms that are related to $v$. We use $\mathcal{R}_v$ to denote the set of all the possible implicit semantic terms that are relevant to $v$.


The implicit semantic meanings generally have the following features.

\noindent{\bf Randomness: } Due to the human nature of the message generation users, the implicit semantic meaning that can be expressed by each user is generally not deterministic but exhibits a certain randomness. We use $p({u_v|v})$ to denote the probability of inferring implicit semantic terms $u_v$ when observing $v$, for $v \subseteq \mathcal{V}$ and $u_v \subseteq \mathcal{R}_v$.  


\noindent{\bf Polysemy: } It is known that different users may infer different meanings when observing the same explicit term, e.g., some users may
infer the TV cartoon character when observing the term ``Tweety", while for others, the term ``Tweety" may mean smartphone App of a social network website. 

\noindent{\bf Inference Rule: }
Recent study suggests that, for each individual user, its inference preference can be characterized by a function that maps the explicit semantics to the possible implicit semantic terms and/or concepts. In this paper, we follow the same setting and, to characterize the randomness of implicit semantic inference process, we define the inference rule of the source user, denoted as $\pi\left( v \right)$, as a function mapping each given explicit semantic term $v$ to a probability distribution of all the possible implicit semantic terms. We assume $\pi\left(v \right)$ is stationary and also $\sum \limits _{u_v \subseteq \mathcal{R}_v}p_{\pi}({u_v | v}) = 1$, where $p_{\pi}({u_v | v})$ is the probability of inferring $u_v$ based on inference rule $\pi$ when observing $v$.
Note that we assume neither source user nor destination user can know the true inference rule of the source messages. The source user can however observe a set of expert message samples, consisting of implicit semantics generated by the source user during the previous communications.
\subsection{Semantic Communication Model}
A general semantic communication model consists of the following key components. 
\begin{itemize}
    \item[(1)] {\bf Semantic Recognizer:} extracts the key explicit semantic terms from the observed messages. For example, if the observed messages are in the form of images or voice and the semantic terms are object labels, it can directly apply existing object detecting algorithms, such as YOLO and wav2letter, to extract the explicit semantics.   

    \item[(2)] {\bf Semantic Encoder:}  converts the extracted semantic information into a form that is suitable for physical channel transmission. In the traditional semantic communication model, the source user has two types of encoders: semantic (source) encoder for minimizing the redundancy in the semantic information and channel encoder for improving the robustness against noisy channel corruption by adding a certain amount of redundancy into the transmitted signals. Recently, the joint source-and-channel encoding has also been considered to implement both types of encoders into a single encoding function, e.g., a DNN, that directly converts the input message into an output signal for physical channel transmission. In this paper, we
    assume the semantic encoder directly converts the semantic information identified by semantic recognizer into the signal to be transmitted to the physical channel. We will discuss in details later in this paper. 


    \item[(3)] {\bf Semantic Decoder:} recovers the complete semantic information at the destination user. In explicit semantic communication, the main objective of the destination user is to recover the explicit semantics identified by the semantic recognizer of the source user. In this paper, we consider the implicit semantic communication, in which the destination user needs to recover both explicit and implicit semantics. Let $\hat \omega = \langle \hat v, \hat u_v \rangle$ be the recovered semantic meaning of the destination user, where $\hat v$ and $\hat u_v$ are the recovered explicit and implicit semantics.  
\end{itemize}

The main design objective of the semantic communication system is to minimize the
semantic distance between the original semantic information and the recovered meaning at the destination user. Let $\Gamma \left( \omega, \hat \omega \right)$ be the semantic distance between $\omega$ and $\hat \omega$.

In this paper, we focus on the implicit semantic communication in which the destination user needs to recover both explicit and implicit semantics. Since implicit semantic cannot be directly identified by semantic recognizer,
the main objective is then to learn an approximated inference rule $\pi_d$ that can minimize the semantic distance between the explicit semantics and possible implicit semantics inferred by the true inference rule of the source user as well as those recovered by the learned inference rule, we formulate the optimization problem of implicit semantic communication as:
\begin{eqnarray}
\min\limits_{\pi_d} \mathbb{E} \left[ \Gamma \left( \langle v, u_v \rangle, \langle \hat v, \hat u_v \rangle \right) \right].
\end{eqnarray}

A straightforward approach for implementing implicit semantic communication is to let the source user infer the implicit semantics from the explicit semantics identified by semantic recognizer, and then transmit both types of semantics to the destination user. This approach however suffers from the following challenges. First, as mentioned earlier, due to randomness and polysemy, implicit semantic information consists of the probability distributions of a set ${\mathcal R}_v$ of possible implicit semantic terms, characterized by a set of floating-point values, each requiring a large number of data packets for semantic information transmission. Second, allowing the source user to perform implicit semantic inference whenever it observes explicit semantics may result in extra delay for information transmission. Moreover, the quality of the recovered implicit semantics may be closely related to the service need of the destination user. Always letting the source user send all the potential implicit information will result in reduced communication efficiency. Finally, the implicit semantics are often generated by the personally preference-related inference rule. Therefore, transmitting implicit semantics will also cause privacy information exposure.

In this paper, we propose a novel adversarial learning-based solution called iSAC for the destination user to learn an inference rule to directly infer implicit semantics based on the explicit semantics sent by the source user. In this way, during communication, the source user only needs to send explicit semantics observed from the source messages and the destination user can automatically recover the intended implicit semantics.    



\section{iSAC Architecture}
In this section, we first define the semantic distance that is suitable to characterize the difference between any given pair of inference rules. We then introduce the iSAC architecture and present the theoretical analysis.
\begin{figure}[t]
\centering
\includegraphics[width=9cm]{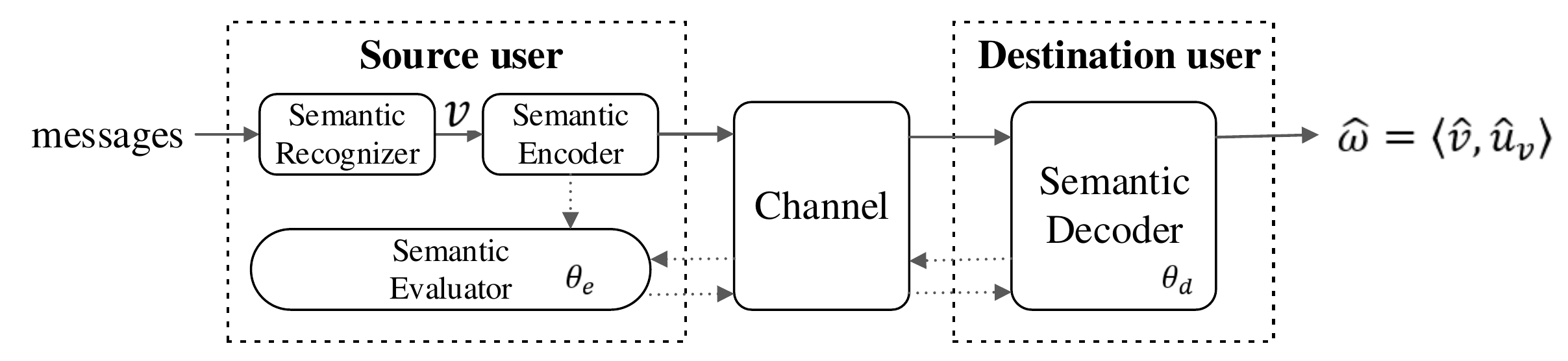}
\caption{\small{Architecture of iSAC.}}
\label{isac}
\end{figure}
\subsection{Semantic Distance}
One of the key differences between the traditional data-oriented communication and the semantic communication solutions is that in the latter, instead of accurately reproducing the signal in its original form, the destination user tries to infer the semantic meaning that matches the real semantics associated with the source messages. This problem becomes more challenging for iSAC because in this system, the destination user needs also to recover implicit semantics that cannot be directly observed in the message received by the source user. Due to the randomness and polysemy of the implicit semantics, an appropriate solution for evaluating the semantic distance is to adopt statistic-based distance measure. In this paper, we focus on recovering the implicit semantics at the destination user. More specifically, the destination user tries to learn the inference rule to infer the correct implicit meaning based on the received signal. Let $\hat{v}$ be the signal decoded by the destination user. The learned inference rule of the destination user $\pi_d$ is a mapping function: $\pi_d: \hat{v} \rightarrow {\cal D}_{\pi_d}(\hat u_v \hspace{-0.06cm} \mid \hspace{-0.06cm} \hat v)$ where ${\cal D}_{\pi_d}(\hat u_v \hspace{-0.06cm} \mid \hspace{-0.06cm} \hat v)$ is the probability distribution of the possible implicit semantics $\hat{u}_v$ when observing $\hat{v}$. Similarly, we define the probability distribution of implicit semantics generated by the true inference rule of the message source when observing explicit semantics $v$ as ${\cal S}_{\pi}(u_v \hspace{-0.1cm} \mid \hspace{-0.1cm} v)$. We use cross entropy, one of the most popular metrics for measuring statistical distances. In this case, the semantic distance between the true implicit semantic and the estimated implicit semantics at the destination user can be written as: 
\begin{equation}
\begin{gathered}
\Gamma \left(\langle v, u_v\rangle, \langle\hat v, \hat u_v\rangle\right)
=\sum_{v \subseteq \mathcal{V}} \left(\mathbb{E}_{u_v \sim \mathcal{S}_{\pi}(u_v |v) }\left[\log p\left(u_v|v \right)\right]\right. \\
\left.+\mathbb{E}_{\hat u_v \sim \mathcal{D}_{\pi_d}(\hat u_v |\hat{v})}\left[\log \left(1-p\left(\hat u_v|\hat{v}\right)\right)\right]\right).
\label{semdis}
\end{gathered}
\end{equation}

Our proposed solution can be directly applied when other statistical distance measures are applied.

\subsection{iSAC Architecture}
In this paper, we propose an adversarial learning-based iSAC architecture, in which
the source user tries to assist in training an inference rule at the destination user, such that the source user only needs to send the explicit semantics observed directly from the original message and the destination user will be able to automatically infer implicit semantics based on the signal sent from the source user. We introduce a new component, the \emph{semantic evaluator}, whose main objective is to compare the implicit semantics estimated by the destination user with a set of expert data samples generated by the true inference rule of the message source. As will be proved later, by introducing the semantic evaluator at the source user, the destination user will be able to learn an inference rule without observing any expert samples. In other words, the expert data samples can only be observed by the semantic evaluator and therefore, the implicit semantic message samples will not be exposed in both training and communication processes.

Let us describe the implementation details of different components of iSAC at the source and destination users as in the following:
\subsubsection{Semantic encoder at the source user}
The main objective of the semantic encoder is to encode the recognized explicit semantics into a suitable form that can be transmitted through physical channels. The source user, for instance, can use the joint source-channel encoder proposed in \cite{oshea2017physicallayerDNN} to encode the explicit semantic information. In this paper, we use the bold font $\boldsymbol{v}$ to denote the encoded version of the explicit embedding sent by the source user.  

\subsubsection{Semantic decoder at the destination user}

The main objective of the semantic decoder is to learn an inference rule that can generate implicit semantic terms based on the signal received from the channel. Suppose the encoded explicit semantic symbol sent by the source user is given by $\boldsymbol v$, we can then write the received explicit semantic symbol received at the destination user as:
\begin{equation}
{\boldsymbol{v}}^d={\boldsymbol H} {\boldsymbol{v}}+{\boldsymbol N},
\label{vd}
\end{equation}
%
where $\boldsymbol H$ is the channel gain and $\boldsymbol N$ is the additive noise received at the semantic decoder. Once received the noisy version of explicit semantics ${\boldsymbol{v}}^d$, the semantic decoder will output the most possible implicit semantics terms related to $\hat v$. We can therefore write the semantic decoder as a mapping function with parameters $\theta_d$, e.g., a graph neural network (GNN) with parameters $\theta_d$, denoted as $\pi_d(\hat{u}_v|\hat v;\theta_d)$.
The output of the semantic decoder is the most possible $\hat{u}_v$ when observing $\boldsymbol{v}^d$. For example, if we adopt a graph softmax based approach for the semantic decoder to estimate the inference probability, the probability for inferring $\hat{u}_v \subseteq {\mathcal{R}_v}$ when observing $\hat v$ can be calculated as
\begin{equation}
\pi_d\left(\hat{u}_v \mid \hat v; \theta_d \right)= \frac{\exp \left(\theta_d^{\top} (\hat{u}_v) \cdot \theta_d (\hat v)\right)}{\sum_{\hat{u}_v \subseteq {\cal R}_v} \exp \left(\theta_d^{\top} (\hat{u}_v) \cdot \theta_d(\hat v)\right)},
\label{pdf_deco}
\end{equation}
where $\theta_d (\hat{u}_v)$ and $\theta_d (\hat v)$ are $\gamma$-dimension representation vector of $\hat{u}_v$ and $\hat v$, respectively.
The semantic decoder can be iteratively trained using SGD, i.e., we can calculate using
\begin{equation}
\theta_d^t = \theta_d^{t-1} - \xi \nabla_{\theta_d} \Gamma_d,
\label{sgd_deco}
\end{equation}
where $\theta_d^t$ are the parameters of semantic decoder at $t$-th iteration, $\xi$ is the learning rate and $\Gamma_d$ is the semantic distance received from the semantic evaluator at the source user.

More specifically, at the beginning of the training phase, the semantic decoder randomly picks up implicit semantic terms in $\mathcal{R}_v$. The selected implicit semantic terms will be feedback to the semantic evaluator at the source user for comparison with the expert message samples. The calculated semantic distance will be sent to the semantic decoder for correcting its inference results and updating its parameters. The decoder will then sample the possible implicit semantic terms in the next iteration according to the updated inference rule $\pi_d\left(\hat{u}_v | \hat v; \theta_d^t \right)$ calculated  by (\ref{pdf_deco}). As will be proved later, the finally trained inference rule at the semantic decoder will be able to approach the true inference rule of the message source. Note that, during the training phase, the semantic decoder only needs to feedback the indices of the inferred implicit semantic terms and therefore the communication overhead of the training phase is limited. Also, after the training phase, the semantic decoder will be able to directly generate the implicit semantics sent by the source user without incurring any feedback or communication overhead.


\subsubsection{Semantic evaluator at the source user}

The evaluator, with the objective of maximizing (\ref{semdis}), evaluates the performance of the decoder in the training phase. 
In other words, the ability of differentiating the semantic distance between the implicit semantics generated by the source user based on $\pi$ and that estimated by the destination user using $\pi_d$ will be enhanced. 
The output $p(u_v\mid v)$ of the semantic evaluator is a probability of the connection existing between $u_v$ and $v$
\begin{equation}
p\left(u_v\mid v \right)=
\frac{1}{1+\exp \left(-{\theta_e}(u_v)^{\top} {\theta_e}(v)\right)},
\label{eva}
\end{equation}
where $\theta_e$ is the parameters of the evaluator, and ${\theta_e}(u_v)$, ${\theta_e}(v)$ are the $\gamma$-dimension representation vectors of semantic terms $u_v$ and $v$.
Any graph representation learning model, such as graph convolutional network (GCN)\cite{semigcn}, can serve to facilitate  the task of embedding. In this paper, we utilize a few graph convolutional layers, which are designed especially for data with graphical structure, to obtain the embeddings of knowledge entities.
The propagation process of the stacking layers can be written as:
\begin{eqnarray}
L^{(0)}=X \  \mbox{and}\  L^{(l+1)}=\sigma(\Phi L^{(l)}W_{l}),
\label{gcnlayer}
\end{eqnarray}
where $\sigma(\cdot)$ is the Sigmoid function, $L^{(l)}$ is the output of layer $l\in \{0, 1, \cdots, m\}$ and $X=[x_1,\cdots ,x_n] \in \mathbb{R}^{n\times\gamma}$ is the initial feature vector of $n$ semantic terms. $\Phi=\tilde{D}^{-\frac{1}{2}} \tilde{A} \tilde{D}^{-\frac{1}{2}}$ is a renormalized Laplacian matrix,  where $\tilde{D}_{i i}=\sum_{j} \tilde{A}_{i j}$, $\tilde{A}=A+I$, $I$ is an $n\times n$ identity matrix and $A$ is the adjacent matrix of the semantic terms. Then the output of the final layer $L^{(m)}$ will be used as $ \theta_e(v)$ to calculate the result of (\ref{eva}).

Suppose $p(u_v | v)$ is differentiable with respect to $\theta_e$. Then both the decoder and evaluator can be iteratively updated using SGD. 
More specifically, the semantic evaluator can be iteratively updated by 
 \begin{equation}
 \theta_e^t = \theta_e^{t-1} - \xi \nabla_{\theta_e} \Gamma_e,
 \label{sgd_eva}
 \end{equation}
where the gradient of (\ref{semdis}) w.r.t. $\theta_e$ is calculated as
%
\begin{equation}
 \nabla_{\theta_e} \Gamma_e =
 \left\{
\begin{array}{l}
\nabla_{\theta_e} \log p\left(u_v\mid v\right), \text { if } u_v \sim {\cal S}_{\pi}(u_v|v);\\
\nabla_{\theta_e}\left(1-\log p\left(\hat{u}_v\mid \hat v\right)\right), \text { if } \hat{u}_v \sim {\cal D}_{\pi_d}(\hat u_v|\hat v).
\label{eva_train}
\end{array}
\right.
\end{equation}

Similarly, the semantic evaluator can calculate the gradient of (\ref{semdis}) w.r.t. $\theta_d$ using the following equation: 
\begin{equation}
 \nabla_{\theta_d} \Gamma_d =
 \nabla_{\theta_d} \sum_{v\subseteq\mathcal{V}} \mathbb{E}_{\hat{u}_v \sim {\cal D}_{\pi_d}(\hat u_v|\hat v)}\left[\log \left(1-p\left(\hat{u}_v\mid \hat v;\theta_e \right)\right)\right].
\label{deco_train}
\end{equation}
Note that, in each iteration, the semantic evaluator at the source user first receives the implicit semantic terms from the semantic decoder and updates its own parameters $\theta_e$ using (\ref{eva_train}). The semantic evaluator will then send the calculated semantic distance value to the semantic decoder to update its parameters $\theta_d$ by (\ref{sgd_deco}) and (\ref{deco_train}).
The detailed algorithm is presented in Algorithm  \ref{Algorithm_JointTraining}.

\footnotesize
\begin{algorithm}[t]
\setlength{\belowdisplayskip}{-1.5cm}
\caption{iSAC Algorithm}
\label{Algorithm_JointTraining}
{\bf Input}: The set of explicit semantic terms $\mathcal{V}$, the set of  relevant implicit semantic terms $\mathcal{R}_v$, initial input feature vectors $x_v$ for semantic term $v$, parameters of evaluator and decoder $\theta_e$ and $\theta_d$, learning rate $\xi$, training iteration $T$\\
{\bf Output}: Inference rule $\pi_d^*$ \\
{\bf Initialization}: Randomly initialize the parameters $\theta_e$ and $\theta_d$\\
{\textbf{For}} $t=0,1,\cdots,T$  {\textbf{do}}
\begin{itemize}
    \item Encoder send the index of $v$ to decoder and evaluator

    \item {\textbf{For}} decoder's steps {\textbf{do}}\\
      \ Decode $\boldsymbol{v}^d \rightarrow \hat{v}$ \\
      \ Samples the most likely connected  $\hat{u}_v$ by (\ref{pdf_deco}) \\
      \ Update $\theta_d$: $\theta_d^t = \theta_d^{t-1} - \xi \nabla_{\theta_d} \Gamma_d$  \\
    {\textbf{end for}}

    \item {\textbf{For}} evaluator's steps {\textbf{do}}\\
    \ Receive negative samples from decoder and derive some positive samples from original message.\\
    \ Calculate the gradient of $\Gamma $ using (\ref{eva_train}) and (\ref{deco_train}). \\
    \ Update $\theta_e$: $\theta_e^t = \theta_e^{t-1} - \xi \nabla_{\theta_e} \Gamma_e$\\
    {\textbf{end for}}
\end{itemize} \leavevmode
\noindent{\textbf{end for}} \\
Calculate $\pi_d^\ast$ by (\ref{pdf_deco})
\end{algorithm}
\normalsize

\subsection{Theoretical Analysis}
Let us now prove that the implicit semantics generated by the learned inference rule utilizing our proposed iSAC can approach that produced by the true inference rule of the source messages.  
\newtheorem{thm}{\bf Theorem}
\begin{thm}\label{thm1}
Suppose the inference rule that dominates the implicit semantics is a stationary process and, in each iteration, the semantic evaluator can always reach its optimal value given by $\theta_e^{\ast}(u_v,v)=\frac{p(u_v| v)}{p(u_v| v)+p(\hat{u}_v | \hat v)}$. 
The probability distribution of implicit semantics generated by the learned inference rule ${\cal D}_{\pi_d}(\hat u_v|\hat v)$ approaches that generated by the true inference rule ${\cal S}_{\pi}(u_v|v)$. 
\end{thm}
\begin{IEEEproof}\label{proof1}
The proof proceeds similarly as the adversarial learning proposed in \cite{goodfellow2020generative}. We summarize the main idea due to the limit of space. 
It can be directly observed that (\ref{semdis})
is convex in ${\cal D}_{\pi_d}(\hat u_v|\hat v)$. By substituting the optimal semantic evaluator into (\ref{semdis}), the resulting function in (\ref{semdis}) is also convex. We can therefore  
apply the gradient descent update for ${\cal D}_{\pi_d}(\hat u_v|\hat v)$ and follow \cite[Theorem 1]{goodfellow2020generative}, with sufficient small learning rate, 
${\cal D}_{\pi_d}(\hat u_v|\hat v)$ can always converge to ${\cal S}_{\pi}(u_v|v)$. 
\end{IEEEproof}



\section{Experimental Results}
\subsection{Dataset and Experimental Setups}
To evaluate the performance of iSAC, we consider two real-world human knowledge datasets: arXiv-GrQc3 and Cora-ML, where arXiv-GrQc3 is a paper citation dataset based on arXiv. It consists of 5,242 vertices, corresponding to the authors, and 14,496 edges, specifying the  collaborations between authors of papers in the general relativity and quantum cosmology categories. Cora-ML is a dataset based on the paper topics in the field of machine learning. It consists of 2,995 vertices, corresponding to feature vectors for each document, and 8,416 edges, specifying the citation links between documents.



All experiments are performed on a Linux workstation (CPU: Intel(R) Xeon(R) CPU E5-2683 v3@ 2.00GHz, GPU: four NVIDIA GeForce GTX 2080Ti (11GB)) and mainly using an open-source Python libraries, Pytorch.

We set two-layer GCN at the semantic evaluator and set the learning rate of the SGD at 0.001. We use semantic decoder defined in (\ref{pdf_deco}) and randomly select the initialized parameters of the semantic decoder based on a uniform distribution. The dimensional size of both representation vectors of the semantic evaluator and decoder are set to 50.  
We randomly select 5\% and 10\% of the links to simulate the expert implicit data samples observed by the semantic evaluator that are connected with a selected number of explicit terms in each dataset.
\subsection{Experiment Results}
\begin{figure}[t]

  \begin{minipage}[t]{0.49\linewidth}
   \centering
   \includegraphics[width=4.7cm]{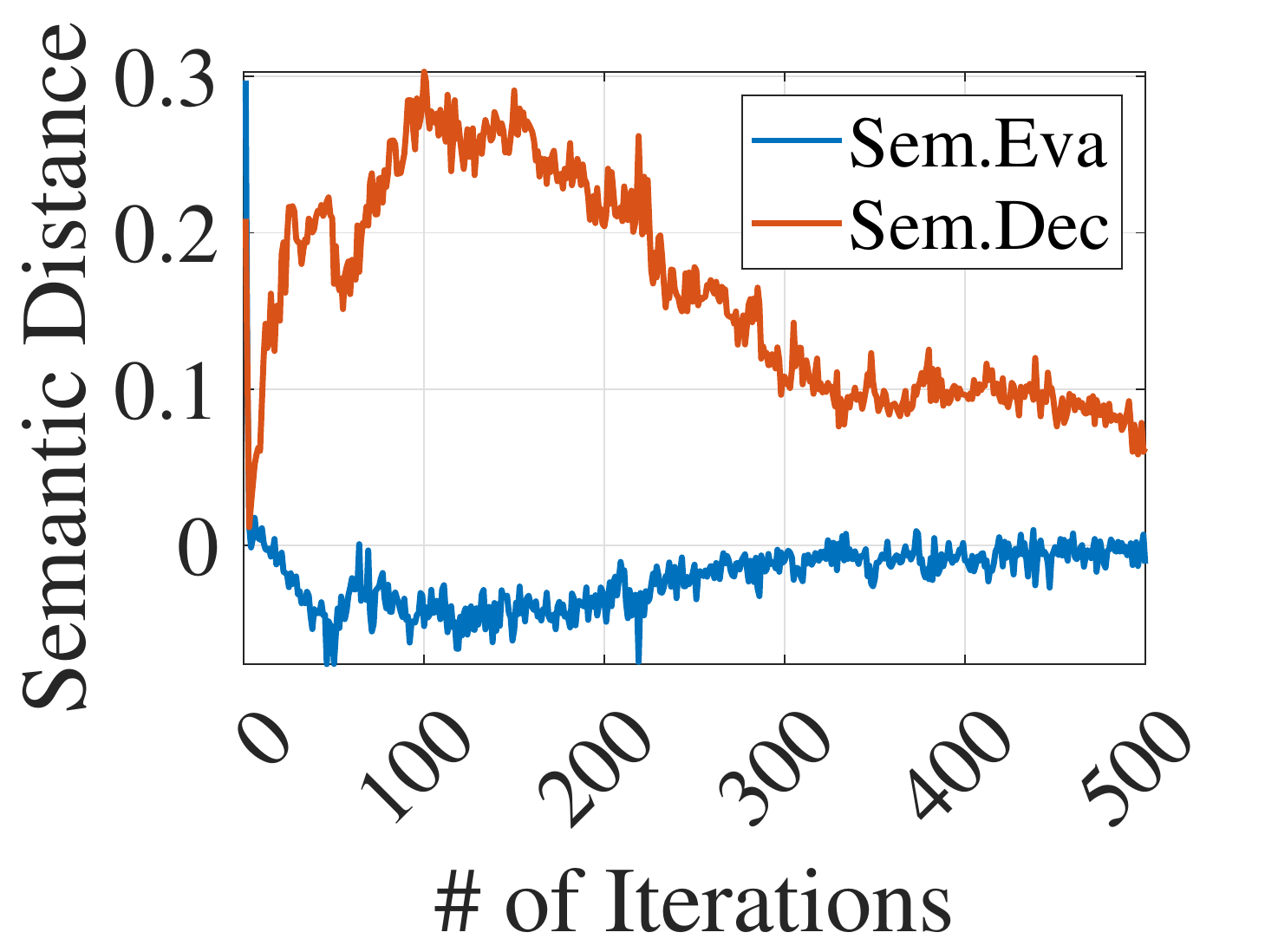}{}
   \caption{\small{Convergence rates of semantic evaluator and decoder. }}
   \label{fig_converge}
  \end{minipage}%
\
  \begin{minipage}[t]{0.49\linewidth}
   \centering
   \includegraphics[width=4.7cm]{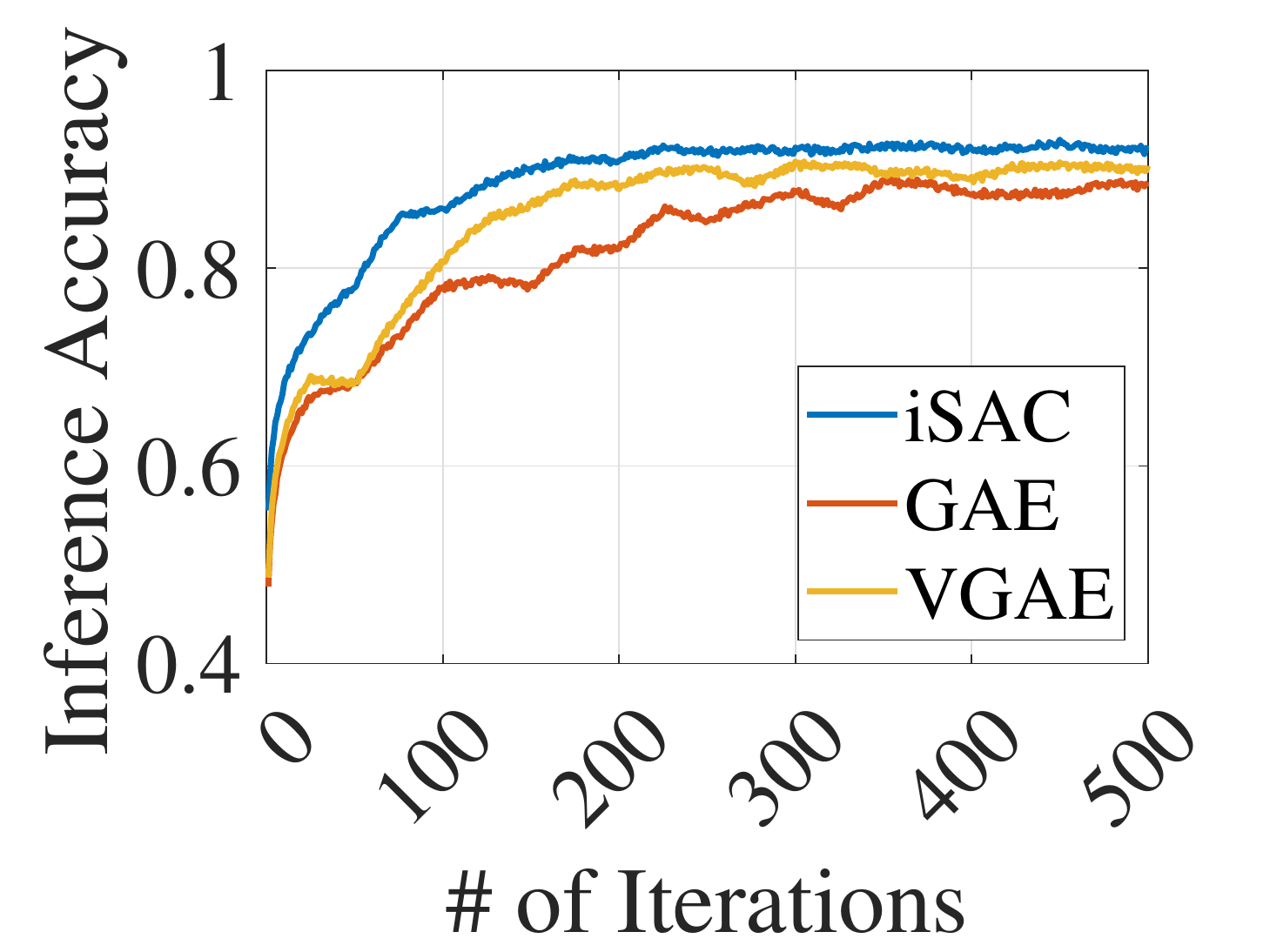}
   \caption{\small{Comparison of the inference accuracy of implicit semantics at the destination user under iSAC, GAE, and VGAE.}}
   \label{fig_inferenceacc}
  \end{minipage}%
\end{figure}


\begin{figure}[t]
\setlength{\abovecaptionskip}{-0.01cm}
\subfigure[]{
  \begin{minipage}[t]{0.45\linewidth}
   \centering
   \includegraphics[width=4.6cm]{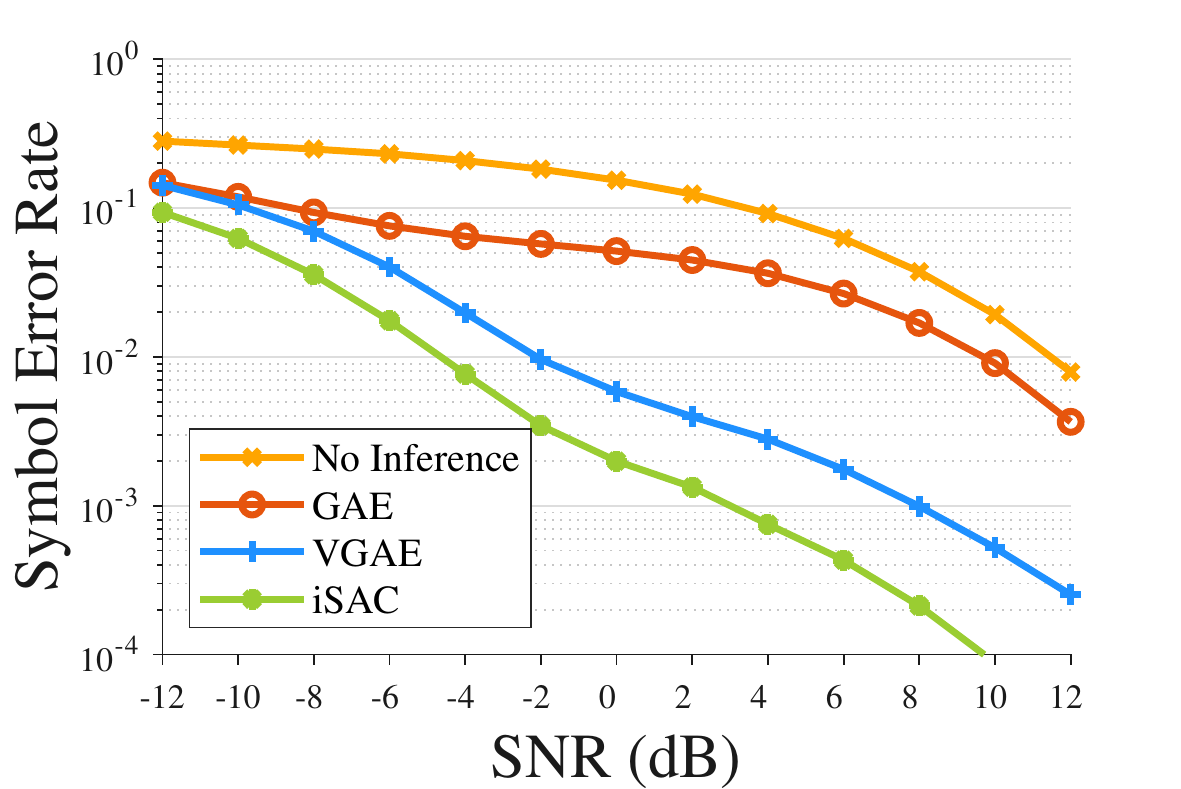}
  \end{minipage}%
}
\  \
\subfigure[]{
  \begin{minipage}[t]{0.45\linewidth}
   \centering
   \includegraphics[width=4.6cm]{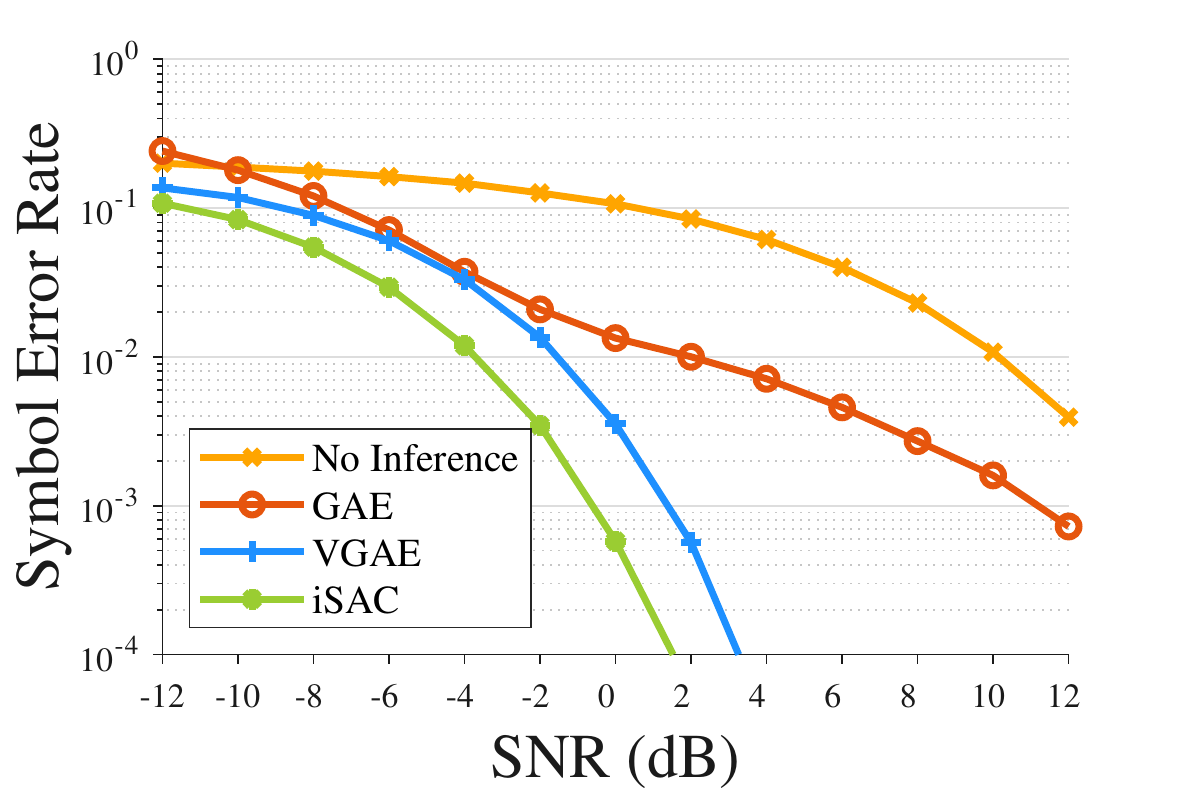}
  \end{minipage}%
}
\caption{\small{Symbol error rate of semantic symbols (entities) when the inference rules learned by iSAC, GAE, and VGAE can be used in semantic symbol recovery, compared to the no inference solution, under datasets (a) arXiv-GrQc3 and (b) Cora-ML.}}

\label{Fig_snr}
\end{figure}

Let us now evaluate the performance of our proposed iSAC. We compare iSAC with the following two solutions implemented to recover implicit semantics as the benchmark.
\begin{itemize}
    \item[(1)] Variational Graph Auto-encoder (VGAE)-based solution: The implicit semantics are first recovered by the semantic recognizer at the source user and then converted into low-dimensional embeddings using a GCN. The converted embeddings will then be sent to the physical channel. The semantic decoder is also a GCN trained to recover the full implicit semantics based on the noisy version of the embeddings sent by the encoder.

    \item[(2)] Graph Auto-encoder (GAE)-based solution: This solution is almost the same as the VGAE-based solution with the only difference that the GCN in the semantic decoder has been replaced as a sigmoid function.
\end{itemize}


Let us first evaluate the convergence performance of iSAC. In Fig. \ref{fig_converge}, we consider dataset arXiv-GrQc3 and  present the loss function, i.e., the semantic distance $\Gamma$, optimized by semantic evaluator and decoder using SGDs.
We can observe that the semantic evaluator and decoder of our proposed iSAC can always converge to each other.

In Fig. \ref{fig_inferenceacc}, we compare the accuracy of the inference rule learned by the semantic decoders of iSAC with that of VGAE and GAE-based solutions. We can observe that our proposed iSAC can always achieve the highest accuracy level for inferring the implicit semantics. In particular, with 100 iterations, VGAE and GAE-based solutions can achieve 80.61\% and 77.79\% inference accuracy levels, respectively. Our proposed iSAC is able to achieve the accuracy of 86.01\%, bringing over 5.40\% and 8.22\% improvements. 

It can be observed that the inference rule learned by the destination user can also be applied to recover semantic information corrupted during the physical channel transmission. In particular, in Fig. \ref{Fig_snr}, we compare the semantic symbol error rate under different inference rules learned by iSAC, VGAE and GAE-based solutions using two different datasets. We can observe that for both datasets, our proposed iSAC always offers the lower symbol error rate, compared to VGAE and GAE-based solutions. In particular, when the SNR of the destination user is 2 dB, iSAC offers 19.69 dB improvements in terms of symbol error rate over the traditional data-oriented communication solutions without any semantic inference. When comparing to the VGAE and GAE-based solutions, the proposed iSAC can offer 4.73 dB and 15.26 dB improvements, respectively. We can also observe that the inference accuracy of all three solutions are better in arXiv-GrQc3, compared to Cora-ML. This is because the semantic terms are more closely linked (with higher degree) in arXiv-GrQc3 than that in Cora-ML. In other words, our proposed inference-rule-based implicit semantic communication solutions are more suitable for message sources consisting of closely linked semantic terms.



\begin{figure}[t]
\subfigure[]{
  \begin{minipage}[t]{0.4\linewidth}
   \centering
   \includegraphics[width=4.3cm]{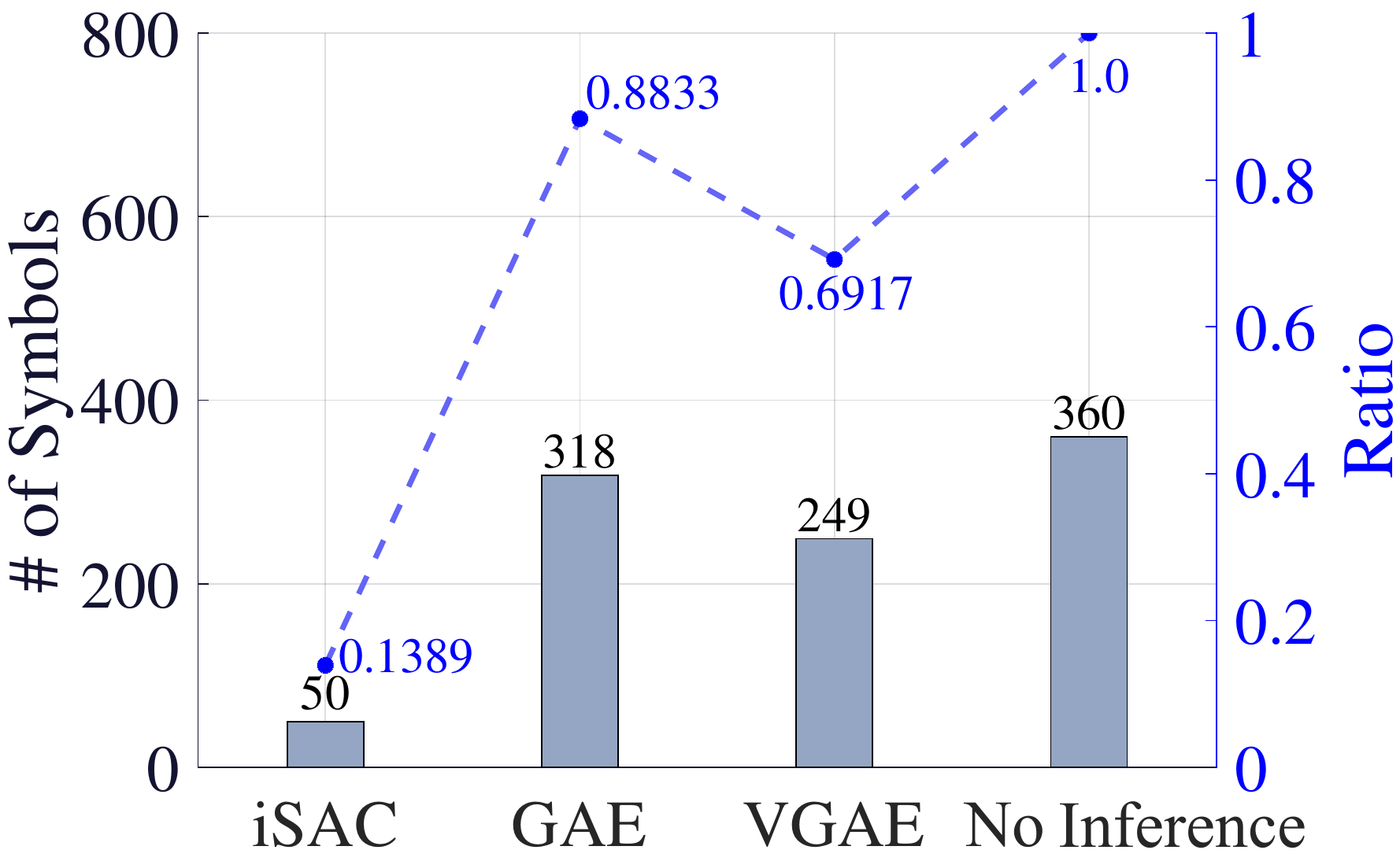}
  \end{minipage}%
}
\  \  \  \  \
\subfigure[]{
  \begin{minipage}[t]{0.4\linewidth}
   \centering
   \includegraphics[width=4.3cm]{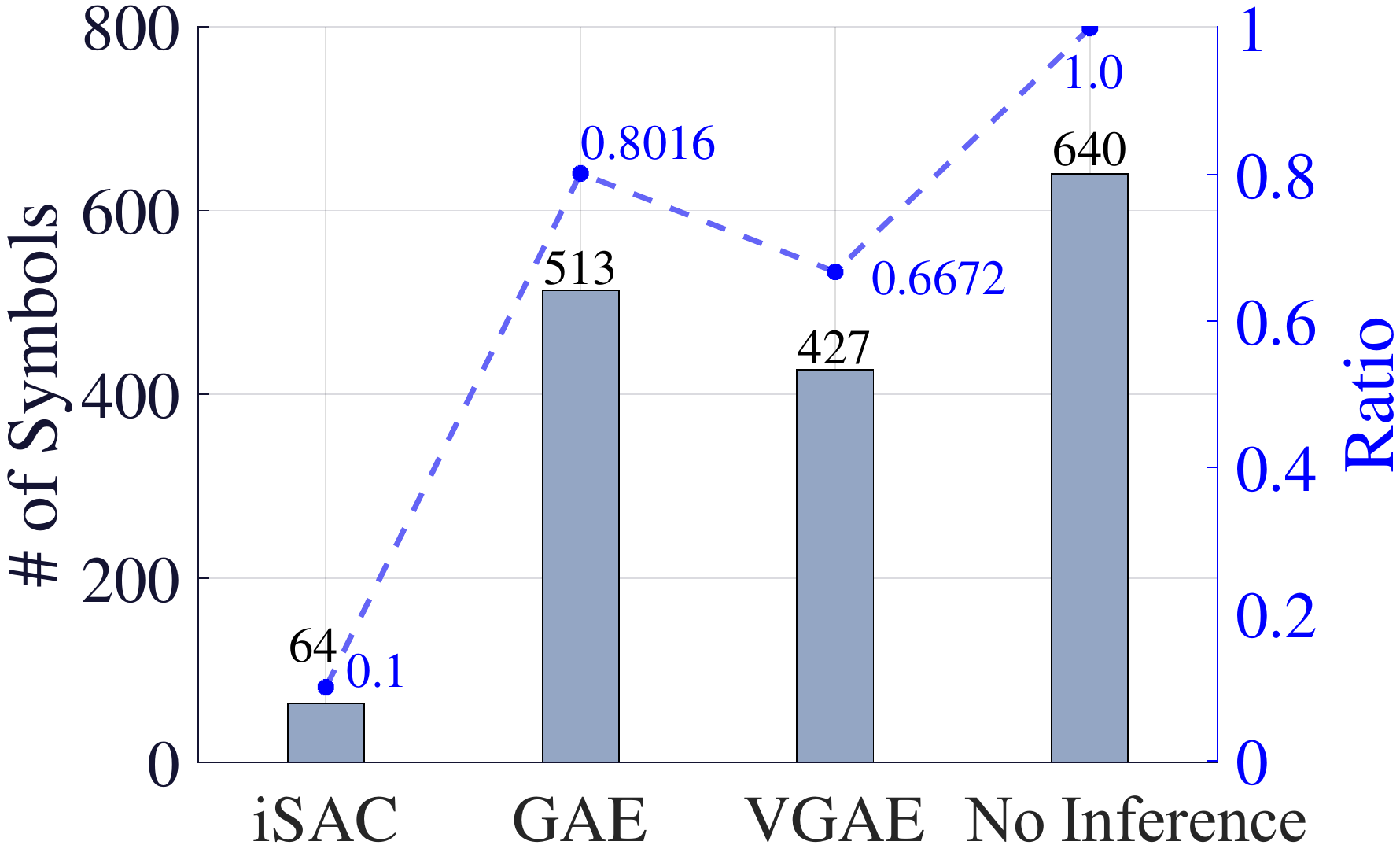}
  \end{minipage}%
}
\caption{\small{Number of required transmitted symbols  for recovering the same amount of implicit semantics at the destination users as well as the corresponding ratios compared to the no inference solution under iSAC, VGAE and GAE-based solutions, under datasets (a) arXiv-GrQc3 and (b) Cora-ML.}}

\label{symbol}
\end{figure}


To evaluate the communication efficiency of iSAC, we compare the required number of fixed dimensional-sized transmitted symbols for recovering the same amount of implicit semantics at the destination user in Fig. \ref{symbol} when implementing iSAC, VGAE and GAE-based solutions in arXiv-GrQc3 and Cora-ML datasets. We can observe that, compared to the non-inferential solution, iSAC, VGAE and GAE-based solutions can achieve over 13.89\%, 88.33\% and 69.17\% improvements in terms of compression rate of transmitted symbols in dataset arXiv-GrQc3 and over 10\%, 80.16\%, and 66.72\% improvements in  dataset Cora-ML. Also, compared to VGAE and GAE-based solutions, iSAC achieves around 84.28\% and 79.92\% reductions in the total number of required transmitted symbols, respectively, in arXiv-GrQc3, and around 87.52\% and 85.01\% reductions, respectively, in Cora-ML. This is because, in our proposed iSAC, the source user only needs to send the explicit (clue) semantics to the destination user. In the VGAE and GAE-based solutions, however, the source user needs to first recover all the implicit semantics and then convert all these semantics into the low-dimensional embedding representations for the physical channel transmission.

\section{Conclusion}
This paper has considered the implicit semantic communication problem in which, instead of sending explicit semantics, hidden relations and closely related semantic terms that cannot be recognized from the source signals need also be delivered to a destination user. We have developed a novel adversarial learning-based iSAC architecture in which the source user tries to assist the destination user to learn an inference rule that can automatically generate implicit semantics based on limited clue information. We have proved that by applying iSAC, the destination user can always learn an inference rule that matches the true inference rule of the source messages. Our experimental results have shown that the proposed iSAC can offer up to 19.69 dB improvement over existing non-inferential communication solutions, in terms of symbol error rate of the destination user.

\section*{Acknowledgment}
Y. Xiao and G. Shi were supported in part by the major key project of Peng Cheng Laboratory under grant PCL2021A12. Y. Xiao was supported in part by the National Natural Science Foundation of China under grant 62071193 and the Key R \& D Program of Hubei Province of China under grants 2021EHB015 and 2020BAA002. G. Shi was supported in part by the National Natural Science Foundation of China under grants 62293483, 61871304, and 61976169. X. Chen was supported in part by the Zhejiang Lab Open Program under Grant 2021LC0AB06. M. Bennis was supported in part by the SNC project under grant No. SCR6GE. H. V. Poor was supported in part by the U.S. National Science Foundation under Grants CCF-1908308 and CNS-2128448.


\bibliographystyle{IEEEtran}
\bibliography{icc2023_lzm}

\begin{thebibliography}{10}
\providecommand{\url}[1]{#1}
\csname url@samestyle\endcsname
\providecommand{\newblock}{\relax}
\providecommand{\bibinfo}[2]{#2}
\providecommand{\BIBentrySTDinterwordspacing}{\spaceskip=0pt\relax}
\providecommand{\BIBentryALTinterwordstretchfactor}{4}
\providecommand{\BIBentryALTinterwordspacing}{\spaceskip=\fontdimen2\font plus
\BIBentryALTinterwordstretchfactor\fontdimen3\font minus
  \fontdimen4\font\relax}
\providecommand{\BIBforeignlanguage}[2]{{%
\expandafter\ifx\csname l@#1\endcsname\relax
\typeout{** WARNING: IEEEtran.bst: No hyphenation pattern has been}%
\typeout{** loaded for the language `#1'. Using the pattern for}%
\typeout{** the default language instead.}%
\else
\language=\csname l@#1\endcsname
\fi
#2}}
\providecommand{\BIBdecl}{\relax}
\BIBdecl

\bibitem{XY2018TactileInternet}
Y.~Xiao and M.~Krunz, ``Distributed optimization for energy-efficient fog
  computing in the {T}actile {I}nternet,'' \emph{IEEE Journal on Selected Areas
  in Communications}, vol.~36, no.~11, pp. 2390--2400, Nov. 2018.

\bibitem{XY2022AdaptiveFog}
------, ``Adaptive{F}og: A modelling and optimization framework for fog
  computing in intelligent transportation systems,'' \emph{IEEE Transactions on
  Mobile Computing}, vol.~21, no.~12, pp. 4187--4200, Dec. 2022.

\bibitem{shi2021semantic}
G.~Shi, Y.~Xiao, Y.~Li, and X.~Xie, ``From semantic communication to
  semantic-aware networking: Model, architecture, and open problems,''
  \emph{IEEE Communications Magazine}, vol.~59, no.~8, pp. 44--50, Aug. 2021.

\bibitem{weaver1949recent}
W.~Weaver, ``Recent contributions to the mathematical theory of
  communication,'' \emph{ETC: A Review of General Semantics}, vol.~10, no.~4,
  pp. 261--281, Sep. 1953.

\bibitem{carnap1952outline}
R.~Carnap and Y.~Bar-Hillel, ``An outline of a theory of semantic
  information,'' Res. Lab. Elctron., Massachusetts Inst. of Technol.,
  Cambridge, MA, RLE Tech. Rep. 247, Oct. 1952.

\bibitem{Santhanavijayan2021ASR}
A.~Santhanavijayan, D.~Naresh~Kumar, and G.~Deepak, ``A semantic-aware strategy
  for automatic speech recognition incorporating deep learning models,'' in
  \emph{Intelligent System Design}.\hskip 1em plus 0.5em minus 0.4em\relax
  Springer, Aug. 2021, vol. 1171, pp. 247--254.

\bibitem{Bao2011TowardsTheorySemanticComm}
J.~Bao, P.~Basu, M.~Dean, C.~Partridge, A.~Swami, W.~Leland, and J.~A. Hendler,
  ``Towards a theory of semantic communication,'' in \emph{Proceedings of the
  IEEE Network Science Workshop}, West Point, NY, Jun. 2011.

\bibitem{xiao2022RateDistortion}
Y.~Xiao, X.~Zhang, Y.~Li, G.~Shi, and T.~Basar, ``Rate-distortion theory for
  strategic semantic communication,'' in \emph{Proceedings of the IEEE
  Information Theory Workshop}, Mumbai, India, Nov. 2022.

\bibitem{Guler2018SCGame}
B.~Guler, A.~Yener, and A.~Swami, ``The semantic communication game,''
  \emph{IEEE Transactions on Cognitive Communications and Networking}, vol.~4,
  no.~4, pp. 787--802, Sep. 2018.

\bibitem{Xie2021LDeepSC}
H.~Xie \emph{et~al.}, ``A lite distributed semantic communication system for
  internet of things,'' \emph{IEEE Journal on Selected Areas in
  Communications}, vol.~39, no.~1, pp. 142--153, Jan. 2021.

\bibitem{Huang2021ImageSamanticCoding}
D.~Huang, X.~Tao, F.~Gao, and J.~Lu, ``Deep learning-based image semantic
  coding for semantic communications,'' in \emph{IEEE GLOBECOM}, Madrid, Spain,
  Dec. 2021.

\bibitem{Weng2021DeepSCS}
Z.~Weng and Z.~Qin, ``Semantic communication systems for speech transmission,''
  \emph{IEEE Journal on Selected Areas in Communications}, vol.~39, no.~8, pp.
  2434--2444, Aug. 2021.

\bibitem{seo2021semantics}
\BIBentryALTinterwordspacing
H.~Seo, J.~Park, M.~Bennis, and M.~Debbah, ``Semantics-native communication
  with contextual reasoning,'' \emph{arXiv}, Aug. 2021. [Online]. Available:
  \url{https://arxiv.org/abs/2108.05681}
\BIBentrySTDinterwordspacing

\bibitem{xiao2022ReasoningOnTheAir}
Y.~Xiao, Y.~Li, G.~Shi, and H.~V. Poor, ``Reasoning on the air: An implicit
  semantic communication architecture,'' in \emph{Proceedings of the IEEE ICC
  Workshop}, Seoul, South Korea, May 2022.

\bibitem{XY2023CollaberativeJSAC}
Y.~Xiao, Z.~Sun, G.~Shi, and D.~Niyato, ``Imitation learning-based implicit
  semantic-aware communication networks: Multi-layer representation and
  collaborative reasoning,'' \emph{IEEE Journal on Selected Areas in
  Communications}, vol.~41, no.~3, Mar. 2023.

\bibitem{Liang2022Lifelong}
J.~Liang, Y.~Xiao, Y.~Li, G.~Shi, and M.~Bennis, ``Life-long learning for
  reasoning-based semantic communication,'' in \emph{Proceedings of the ICC
  Workshops}, Seoul, South Korea, 2022.

\bibitem{oshea2017physicallayerDNN}
T.~O’shea and J.~Hoydis, ``An introduction to deep learning for the physical
  layer,'' \emph{IEEE Transactions on Cognitive Communications and Networking},
  vol.~3, no.~4, pp. 563--575, 2017.

\bibitem{semigcn}
T.~N. Kipf and M.~Welling, ``Semi-supervised classification with graph
  convolutional networks,'' in \emph{Proceedings of the International
  Conference on Learning Representations}, Toulon, France, Apr. 2017.

\bibitem{goodfellow2020generative}
I.~Goodfellow \emph{et~al.}, ``Generative adversarial nets,'' in \emph{Proc. of
  the NIPS}, vol.~27, Montreal, Canada, Dec. 2014.

\end{thebibliography}
\end{document}